\documentclass[10pt,twocolumn,letterpaper]{article}

\usepackage{wacv}
\usepackage{graphicx}
\usepackage{times}
\usepackage{epsfig}
\usepackage{amsmath}
\usepackage{amssymb}
\wacvfinalcopy % *** Uncomment this line for the final submission

\def\wacvPaperID{0431} % *** Enter the wacv Paper ID here

\ifwacvfinal\fi
\begin{document}

%%%%%%%%% TITLE
\title{Sensor Adaptation for Improved Semantic Segmentation of Overhead Imagery}

% Authors at the same institution
%\author{First Author \hspace{2cm} Second Author \\
%Institution1\\
%{\tt\small firstauthor@i1.org}
%}
% Authors at different institutions
\author{Marc Bosch, Gordon A. Christie, Christopher M. Gifford \\
The Johns Hopkins University - Applied Physics Laboratory\\
{\tt\small Marc.Bosch.Ruiz@jhuapl.edu}
%\tt\small Gordon.Christie@jhuapl.edu\\
%\tt\small Christopher.Gifford@jhuaple.edu}
}
%\and
%Gordon A. Christie \\
%The Johns Hopkins University - Applied Physics Laboratory\\
%{\tt\small Gordon.Christie@jhuapl.edu}
%\and
%Christopher M. Gifford \\
%The Johns Hopkins University - Applied Physics Laboratory\\
%{\tt\small Christopher.Gifford@jhuaple.edu}
%}
%}

\maketitle
\ifwacvfinal\thispagestyle{empty}\fi

%%%%%%%%% ABSTRACT
\begin{abstract}
    Semantic segmentation is a powerful method to facilitate visual scene understanding. Each pixel is assigned a label according to a pre-defined list of object classes and semantic entities. This becomes very useful as a means to summarize large scale overhead imagery. In this paper we present our work on semantic segmentation with applications to overhead imagery. We propose an algorithm that builds and extends upon the DeepLab framework to be able to refine and resolve small objects (relative to the image size) such as vehicles. We have also investigated sensor adaptation as a means to augment available training data to be able to reduce some of the shortcomings of neural networks when deployed in new environments and to new sensors. We report results on several datasets and compare performance with other state-of-the-art architectures.
\end{abstract}

%%%%%%%%% BODY TEXT
\section{Introduction}

\begin{figure}[h]
\begin{center}
%\fbox{\rule{0pt}{2in} \rule{.9\linewidth}{0pt}}

	 \includegraphics[scale=0.37]{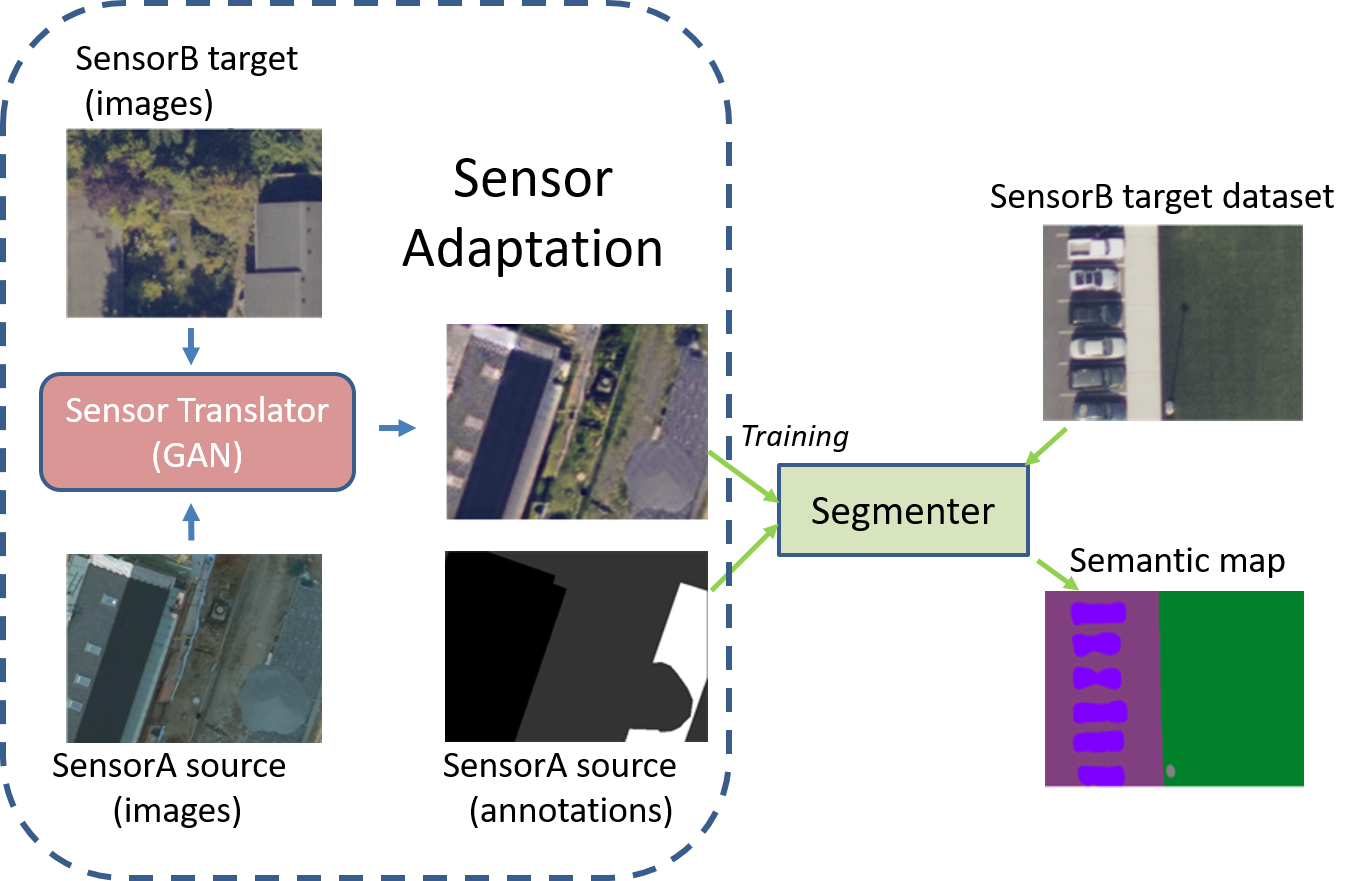}
   \caption{Sensor adaptation block diagram (translator + segmenter). Proposed semantic segmentation system to adapt available training data to a new, unknown sensor.}
\end{center}
\label{fig:diagram}
\end{figure}

In recent years, semantic segmentation has received a lot of attention by the computer vision community. It aims to assign a class label to each pixel in the image by performing fine-grained inference. Semantic segmentation has become a stepping stone towards full scene understanding. It is a key enabler for certain applications, including self-driving vehicles, human-computer interaction, 3D semantic reconstruction, image/video editing, and surveillance tasks.\\
\begin{figure}[h]
%\begin{center}
%\fbox{\rule{0pt}{2in} \rule{.9\linewidth}{0pt}}
%\end{center}
	 \includegraphics[scale=0.31]{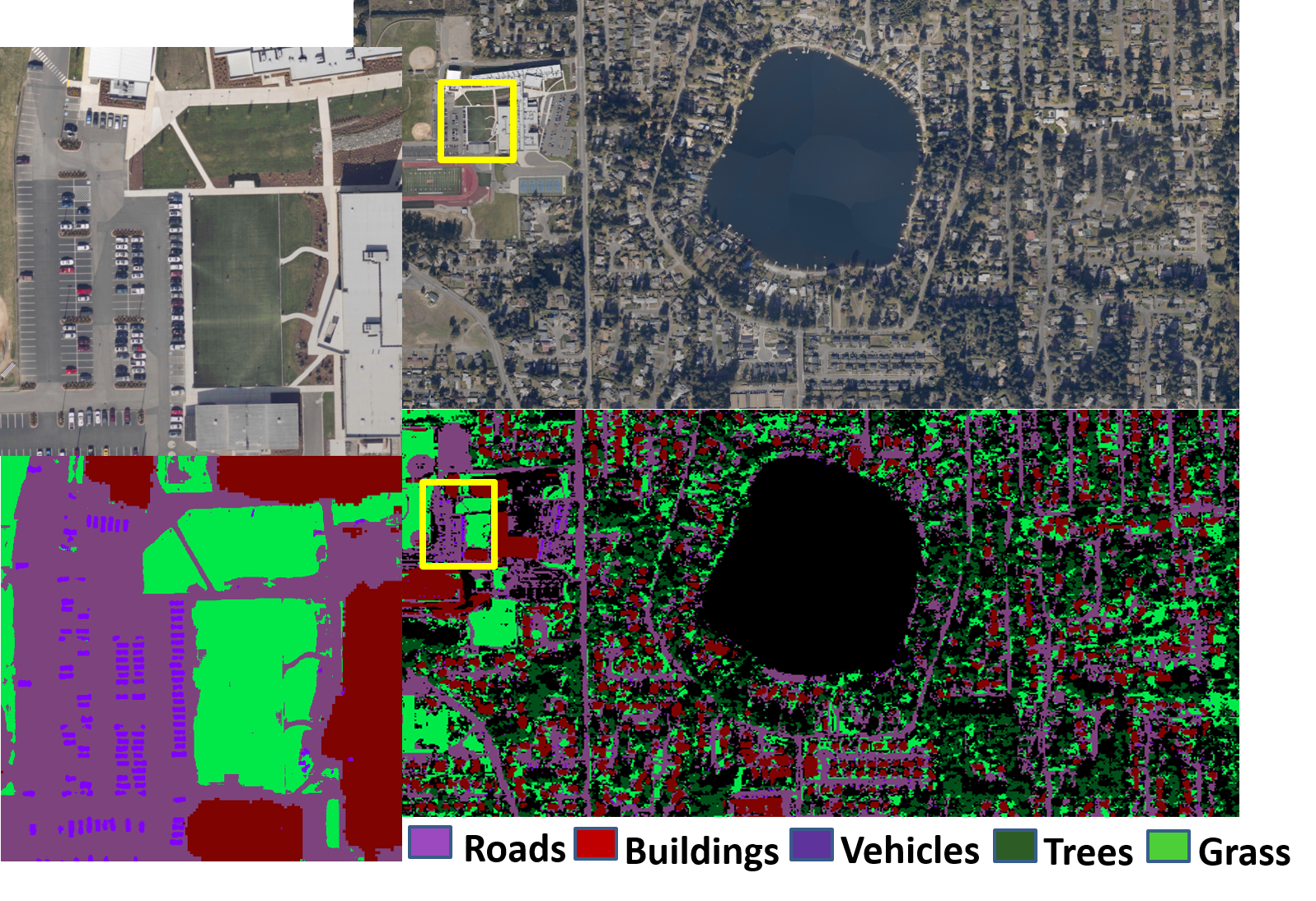}
   \caption{Example segmentation results using the proposed approach for overhead imagery.}
\label{fig:mosaic}
\end{figure}
In the last few years, there have been many proposed architectures based on convolutional neural networks (CNNs). These include SegNet~\cite{segnet}, RefineNet~\cite{refinenet}, PSPNet~\cite{pspnet}, DeepLab and its derivatives~\cite{deeplabv1, deeplabv2, deeplabv3}, among others. In particular, DeepLab has shown very promising results. However, we have observed that this architecture often fails to provide accurate/clear object boundaries. This has also been reported by others in~\cite{deeplabv3plus}. One of the latest flavors of DeepLab includes atrous convolutions in the backbone feature extractor part of the network. This, and other filtering operations like pooling, often causes the network to miss certain information about natural edges. In overhead imagery, this is critical since many objects are closely-spaced. Thus there is a need to obtain more refined segmentation maps.\\
In this work we propose to expand the DeepLab framework by extending it with a refinement module to produce segmentations with less over/under-segmentation outputs. In particular, we take the original DeepLabv3 backbone, which involves a ResNet101 architecture with atrous convolutions, and feed it to a multi-resolution module to progressively recover information as resolution is restored. Figure~\ref{fig:mosaic} shows an example segmentation product of the proposed model for an overhead image.\\
In addition to architecture changes, the other primary factor that influences the performance of neural networks is training data. As in other domains, performance on overhead imagery is sensitive to the training data used. In particular, how well it represents the test data. There are factors such as sensor characteristics, viewing geometry, resolution, and image modality that influence the performance across data sources due to data inconsistencies between the training and testing sets. Further, data annotation for overhead imagery is very costly and time consuming due to the large amount of objects in scenes and their large size. Therefore, it is critical to be able to re-use previous labeling efforts in as many contexts as possible. To this end, we set to examine how to ``translate'' our training data from a given source sensor to look more like the target sensor. The goal is to transfer visual characteristics from source sensor imagery to target sensor imagery. Generative Adversarial Networks (GANs) have been shown to accomplish this goal. Frameworks like pix2pix~\cite{pix2pix} and derivative networks~\cite{pix2pixHD, starGAN, comboGAN} can successfully translate images from one domain to another. In this work, we have integrated a ``translator'' module into the image segmentation task to re-purpose training data to other sensors and platforms. Figure~\ref{fig:diagram} shows the proposed conceptual framework. We have experimented with several simulated sensor types to quantify the gains of such a module. We report our results on several overhead test datasets.\\
%-------------------------------------------------------------------------
%\subsection{Contributions}
In this paper we describe our proposed algorithm and the integration of sensor adaptation for the task of semantic segmentation for overhead imagery. %, where we don't seek the objective quality gains but informantion discovery for enhanced capabilities.
Our main contributions can be summarized as follows:
\begin{itemize}
\item Proposed a semantic segmentation architecture using DeepLabv3 as a backbone and a multi-resolution processing pipeline. 
\item Proposed a two-network system to perform sensor adaptation for improved segmentation results. 
\item Evaluated several simulated sensor models to measure the added value of a sensor adaptation block.
\end{itemize}
The remainder of the paper is organized as follows: In Section 2 we review our proposed semantic segmentation algorithm. In section 3 we describe the sensor adaptation module. In section 4 we show our experimental results on overhead imagery. Finally, we conclude the paper by offering some remarks from our experimental observations.
%-------------------------------------------------------------------------\section{Formatting your paper}
\section{Semantic Segmentation Model}
In recent years, many semantic segmentation models have been published. Many follow a similar structure with a backbone deep neural network to extract features at one or more resolution scales and a collection of high-level layers to combine such features as powerful tools for prediction tasks. One of the most successful models, known as DeepLab, was introduced in~\cite{deeplabv3} and expanded upon~\cite{deeplabv1, deeplabv2}.
\subsection{DeepLab} 
The DeepLab framework leverages dilated convolutions and spatial pyramid pooling layers in the context of semantic segmentation. The baseline model consists of a modified ResNet architecture, where some convolution operations are replaced by upsampled filters or atrous convolutions allowing control of the resolution/scale at which features are extracted. In addition, having a tunable parameter or \textit{rate} to control the sample spacing of the kernel allows the receptive field to change without adding more parameters into the model. Another key element of DeepLab is the addition of the atrous spatial pyramid pooling (ASPP) module that targets multi-scale processing by passing the feature maps to four parallel atrous convolutional layers with different rates. Having these layers allows the model to encode multi-scale context. 
Through experimentation, one of the shortcomings of this architecture is that results are not refined, or in other words, do not always coincide with natural edges in the scene, creating situations of over/under-segmentation. %Conditional Random Fields (CRFs) are added after the final DCNN layer to obtain refinement. Also, image-level features that encode context are added into the model as one of the ASPP layers to counter these limitations. 
\subsection{Refined DeepLab}
\begin{figure}[h]
	 \includegraphics[scale=0.40]{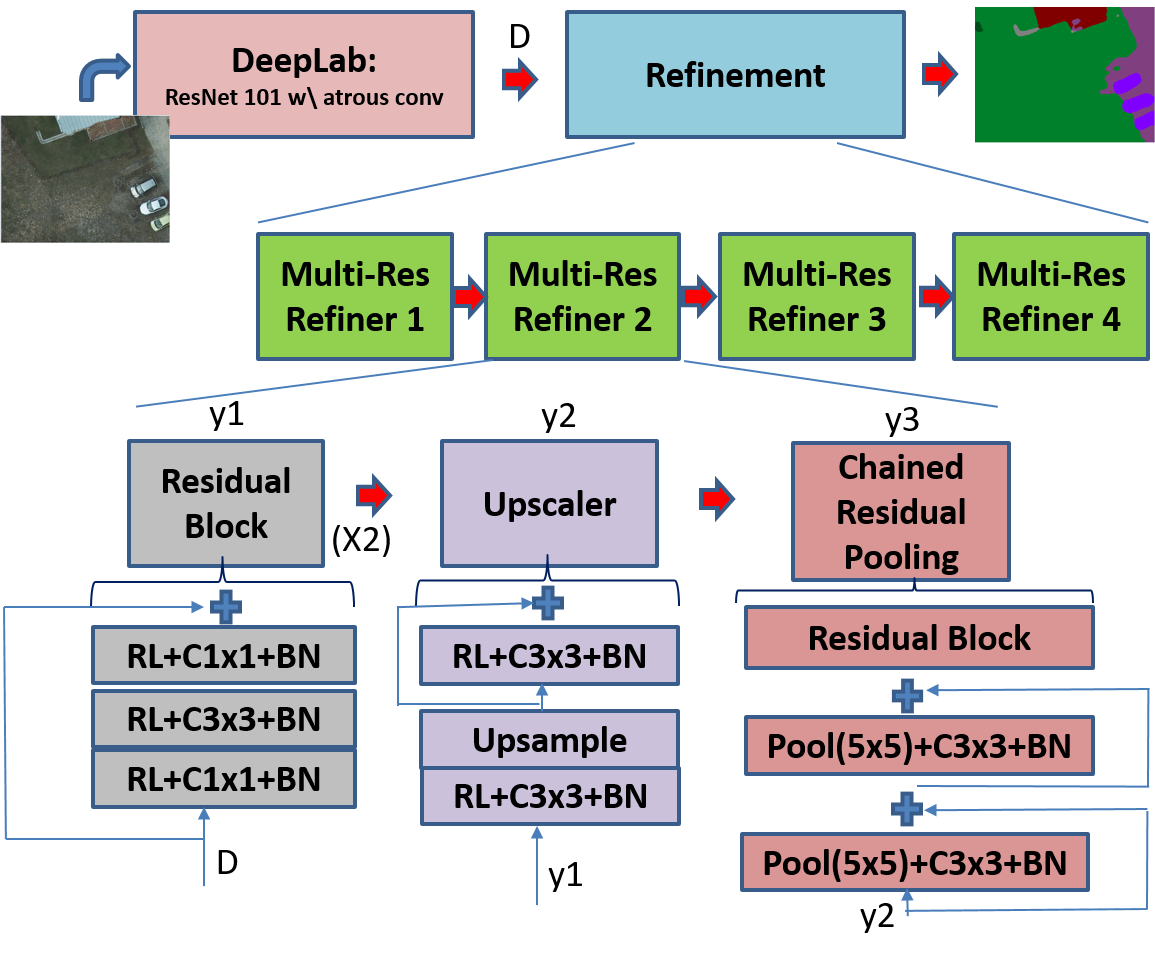}
   \caption{Proposed architecture for overhead imagery semantic segmentation.}
\label{fig:diagram2}
\end{figure}

In overhead imagery, due to the small size of some of the objects of interest, accurate boundaries and object separation is critical to perform analytics and perceptual tasks. In this work we adapted the DeepLabV3 architecture and added a refinement block with the goal of obtaining segmentation masks that are more aligned with the natural edges of the scene.
Our system consists of replacing the ASPP layers and adding a sequence of multi-scale refinement blocks that work on progressively higher resolution feature maps. Each multi-scale refinement modules consist of residual blocks, upscaler (2x factor), and a chained residual pooling similar to~\cite{refinenet}. As mentioned earlier, this configuration is repeated until the full native image resolution is achieved. Figure~\ref{fig:diagram2} shows the block diagram of our modified DeepLabv3 architecture.
The residual blocks consist of a sequence of three ReLU-Conv-BN basic units, with the first and last Conv layers using ($1 \times 1$) kernels, and the convolutional block in the middle being a ($3\times 3$) kernel. The resulting residual, $R_i(x)$, is added to the input, $x$, to obtain the output of the residual block. This operation is repeated one more time, resulting in the final output $y_1=R_2(x+R_1(x))+x+R_1(x)$.\\
After the residual blocks, the output $y_1$ is passed into the upscaler, which consists of two ReLU-Conv-BN units and an upsample operation in between that reverses the effect of a max-pooling by creating a $n\times n$ cell for each value. $n$ is the scale factor desired. The output of the upscaler is denoted as $y_2$ in Figure~\ref{fig:diagram2}. We investigated two flavors of the upsampler: padding zeros and direct copy. Zero-pad fills the $n\times n$ cell with zeros, while direct copy assigns the input value to all positions in the cell. Zero-pad: given a cell in a feature map with value equal to $[0.3]$, the result of the upsample operation is $[0.3\ 0;0\ 0]$, given $n=2$. The direct copy results in $[0.3\ 0.3;0.3\ 0.3]$. In this work we set $n=2$ for all of our experiments.
After the scaler, the final block in each refinement module is a modified chained residual pooling. It consists of the max-pooling+Conv($3\times 3$)+BN sequence, this structure is repeated twice followed by one last residual block (see Figure~\ref{fig:diagram2} for more details). The resulting output, $y_3$ is fed into the next multi-res refiner. Our network consists of a total of 4 refinement levels.

%------------------------------------------------------------------------
\section{Source Adaptation for Overhead Imagery}
Neural network performance applied to overhead imagery often produces underwhelming results compared to ground-based imagery. Even though viewpoints remain relatively consistent in overhead imagery, many images are collected off-nadir (oblique look angle). Other factors that degrade algorithm efficiency are ground sampling distance (i.e., GSD, or image resolution), time of day/year, viewing geometry, and geographical location, among others. Through experimentation, however, we have observed that the most critical contributing factor for loss in performance is due to training on imagery from one sensor and applying the trained model to images of a different sensor. One of our goals is to design semantic segmentation algorithms that can be deployed in new environments, regardless of the acquisition sensor, while incurring minimal accuracy degradation.\\
We tackled this problem by adding a secondary network that acts as a ``translator'' between the source sensor (sensor of which we have training data/labels available) and our target sensor by which test datasets are acquired. Intuitively, by training a segmenter with data that looks more similar to data to be used at test time we should reduce degraded network performance in new environments. Our goal is then to close the gap between the visual appearance of different sets of sensors for improved overall results. The proposed system with sensor adaptation (SA) added to the semantic segmentation network is shown in Figure~\ref{fig:diagram}. To the best of our knowledge, this is the first work that addresses this problem for overhead imagery.\\
Inspired by recent success of domain adaptation frameworks that focus on translating images from one domain to another, as in~\cite{pix2pix,pix2pixHD}, we used a cycle-consistent adversarial framework known as \textit{CycleGAN}~\cite{cyclegan} to work with the overhead imagery. We explored several loss function configurations to adapt it to our problem.\\
We investigated adding a feature loss to preserve the content. The goal is to produce an image from the source sensor as if the target sensor had acquired it by solving the following problem:
\begin{equation}
I_{o}=amin_{I} (\alpha Loss_{adv}+\lambda (\beta Loss_{cycle}+(1-\beta) Loss_{fm}))
\end{equation}
where $Loss_{adv}$ is the ``vanilla'' GAN adversarial loss, defined as:
\begin{equation}
Loss_{adv}=E_{z\sim p_{model}(z)}(log(1-D_{\psi_D}(G_{\psi_G(z)})))
\end{equation}
where G, the \textit{generator}, generates images G(z) that look similar to images from the target sensor, and the \textit{discriminator}, D, performs a binary (fake vs. real) classification on images from the target sensor and those generated by G.\\
The next term in the objective function is the $Loss_{cycle}$. It enforces cycle consistency between source and target images so that the learned image translation mapping should be able to take a source sensor image, translate it into a target sensor image, and bring it back to the source sensor's content and style using an auxiliary generator $F$. In other words, it satisfies backward consistency by adding the target-to-source content loss:
\begin{equation}
\begin{split}
Loss_{cycle} & =E_{z\sim p_{model}(z)}(|F(G(z))-z|)+\\
 & +E_{x\sim p_{model}(x)}(|G(F(x))-x|)
\end{split}
\end{equation}
Following~\cite{gatys, bosch, johnson}, the $Loss_{fm}$ attempts to describe the visual attribute loss. Using a pre-trained VGG19 on ImageNet, feature vectors of generated and target images are extracted and compared according to:
\begin{equation}
Loss_{fm}=|T_{\theta_{vgg19}(target)}-T_{\theta_{vgg19}(I_{o})}|
\end{equation}
In early experiments we observed that both cycle loss and adversarial loss had the biggest contribution in generating high fidelity data, given the added overhead of the VGG19 network. In addition, we experimented with the style loss proposed in~\cite{gatys}, but we did not observe any qualitative or quantitative improvement. 

\subsection{Implementation and Training Details}
The segmenter of the proposed system uses DeepLabv3's backbone, which consists of ResNet101 with atrous convolutions and batch normalization (BN) added to the network~\cite{deeplabv3}. We replaced the ASPP module with our refinement module. The refinement module consists of four modules composed of residual blocks+upscaler (x2)+chained residual pooling block. The output of the final block is passed to the $1\times 1$ final convolution that generates the final logits. All convolutional layers consist of 256 filters. Depending on the layer, kernel size is either set to $1\times 1$ or $3\times 3$. See Figure~\ref{fig:diagram2} for more details. The max-pooling unit in the chained residual pooling is a $5\times 5$ operation, following~\cite{refinenet}. In all our experiments, the network is trained for 10000 steps using a batch size of 4, and the PASCAL dataset is used to pre-train the backbone ResNet101.\\
The translator's generator consists of 64 filters in its convolutional layers, as does the discriminator network ~\cite{cyclegan}. We set the parameter controlling the contribution of the adversarial loss $\alpha$ to 1, while $\lambda$ is fixed to 10 to guide the network toward perceptual convergence. The translator is trained for 100 epochs in all of our experiments.
%------------------------------------------------------------------------
\section{Evaluation}
In this section we describe our experiments and present results for semantic segmentation of overhead imagery. We compare several methods using objective quality metrics such as mean intersection over union (mIoU), and F1 score, as well as show visual results of the proposed method. The mIoU measures ratio between the intersecting area of the predicted segmentation mask and the ground truth mask, and the area determined by the union of both masks. F1 score is defined as the harmonic mean of both precision and recall. We also conducted studies on the contribution of sensor adaptation for segmentation tasks. In this case the same train/test split was used to both train and evaluate the segmenter and the source adaptation GAN.
%------------------------------------------------------------------------
\subsection{Overhead Imagery Datasets}
We have conducted experiments on several public overhead imagery datasets. These include the ISPRS 2D Labeling dataset~\cite{isprs}, USSOCOM Urban 3D Challenge dataset ~\cite{ussocom}, as well as other imagery from other airborne RGB sensors.\\% See Figure~\ref{fig:dataset} for some examples.\\
The \textit{ISPRS} dataset was collected as part of the 2D Semantic Labeling Contest conducted by the International Society for Photogrammetry and Remote Sensing (ISPRS)~\cite{isprs}. In this work we have used the Potsdam collection, which includes aerial images from the historical city with large building blocks, narrow streets, and dense settlement structures.
The Urban 3D Challenge dataset was released in 2017 as part of the large scale building segmentation challenge~\cite{ussocom}. In this dataset satellite imagery is available with the corresponding binary masks for building annotations.
In our experiments we focused on segmenting the following categories: buildings, vehicles, roads, and (low/high) vegetation. In the case of 3D Urban Challenge dataset, we focused on the buildings category.
Although depth information was available through digital surface models (DSMs) and digital terrain models (DTMs), we have only used the RGB color information in this work.
%\begin{figure}[h]
%	 \includegraphics[scale=0.40]{./figures/dataset.eps}
%   \caption{Example samples used in our overhead imagery experiments.}
%\label{fig:dataset}
%\end{figure}

%------------------------------------------------------------------------
\subsection{Refined DeepLab Model Results}
\begin{figure}[h]
\begin{center}
%\fbox{\rule{0pt}{2in} \rule{.9\linewidth}{0pt}}
	 \includegraphics[scale=0.45]{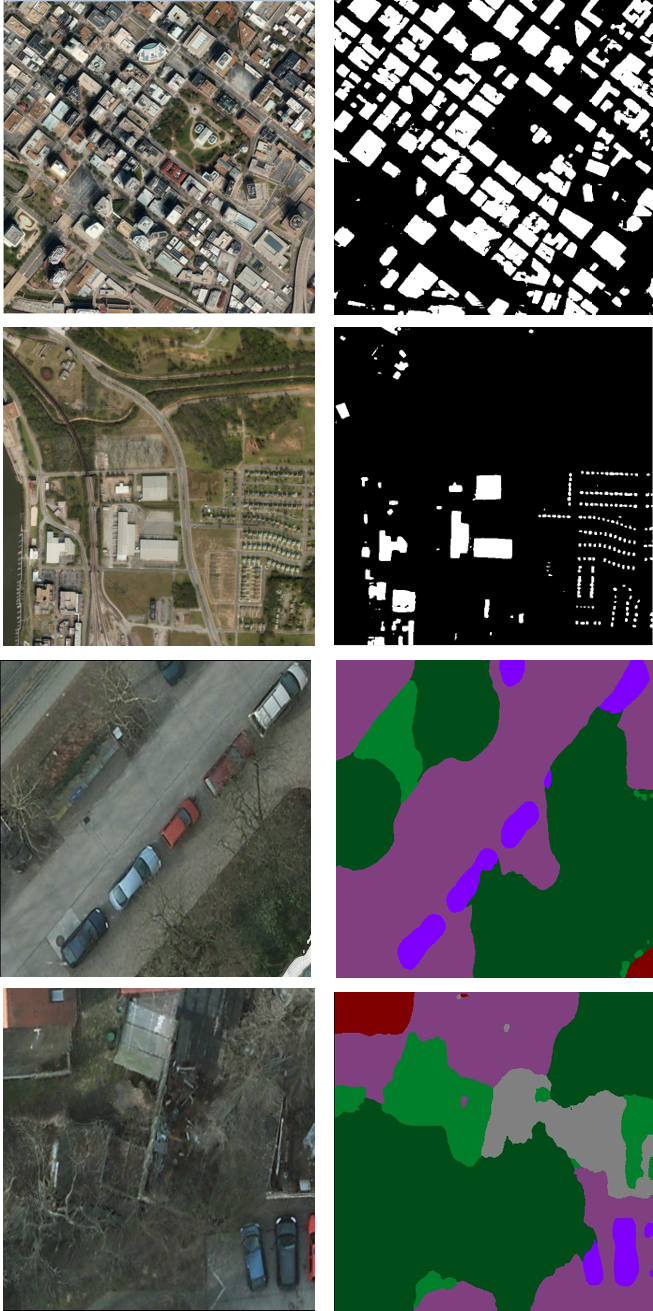}
   \caption{Segmentation results. Top two rows show building segmentation results from USSOCOM Urban Challenge dataset. Bottom two rows show examples from the ISPRS 2D Semantic Labeling Contest.}
\label{fig:examples}
\end{center}
\end{figure}

\begin{figure}[h]
\begin{center}
%\fbox{\rule{0pt}{2in} \rule{.9\linewidth}{0pt}}
%\end{center}
	 \includegraphics[scale=0.45]{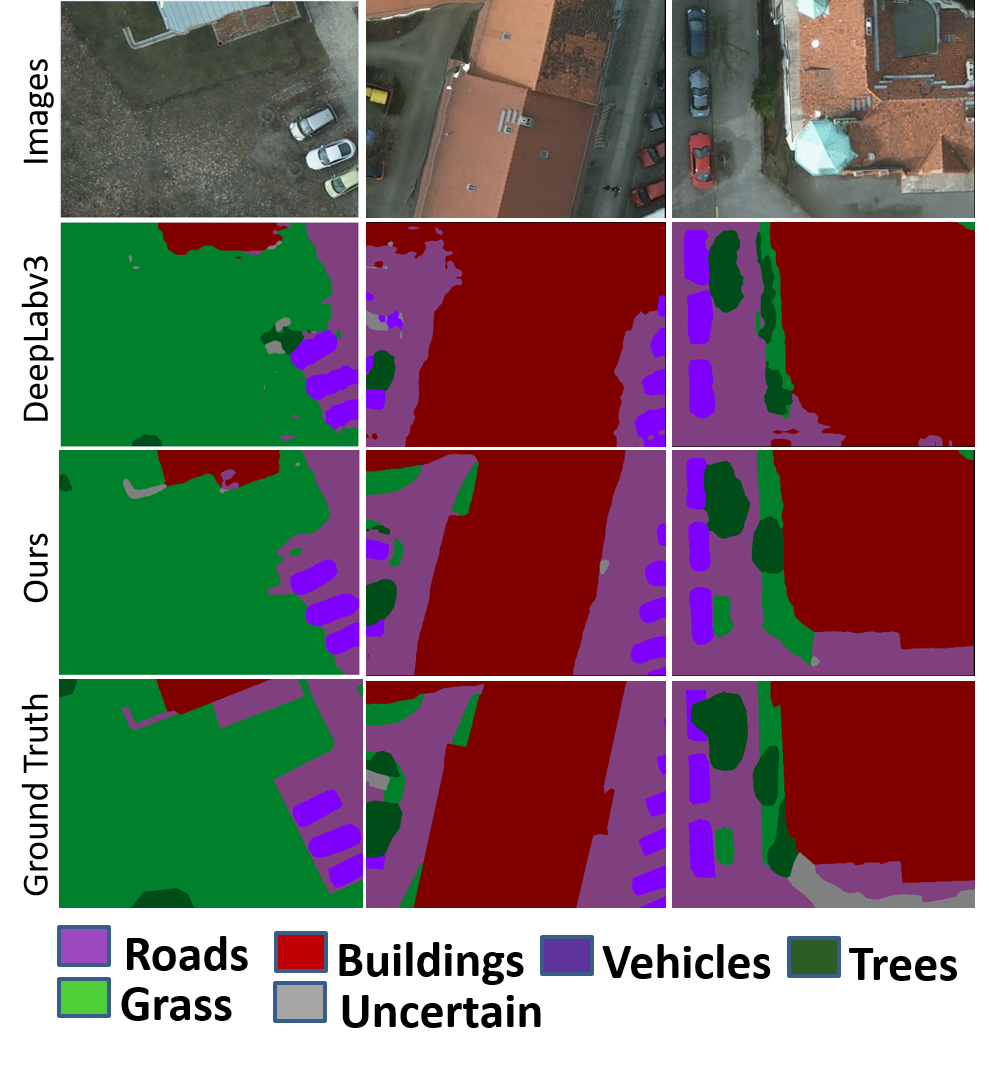}
   \caption{Comparisons between baseline DeepLab3 and our work. First row shows original images, second row shows results of applying DeepLab3/ResNet101, third row shows our proposed method, and last row shows the ground truth annotations.}
\label{fig:dlvstest}
\end{center}
\end{figure}

We have conducted our experiments on several overhead images, as described in the previous section. We have focused on the improvements of the refinement module over the baseline DeepLabv3 algorithm. The first and second rows of Table~\ref{tab:comp} show the performance gain in terms of mIoU and F1-score for all classes.
The first and second rows of Table~\ref{tab:compclass} show metric comparison between baseline and our method for individual classes (vehicles, buildings, and roads).
In addition, we have included qualitative results of the proposed model. Figure~\ref{fig:examples} shows several examples for both building-only segmentation using the Urban 3D Challenge dataset, and multi-class segmentation using the 2D semantic labeling ISPRS dataset. Figure~\ref{fig:dlvstest} shows visual comparisons between baseline DeepLabv3 and our work.

\subsection{Sensor Adaptation Results}
In this section we describe the sensor adaptation (SA) experiments. In this case we have focused the comparisons on building segmentation. We compared performance of the proposed segmenter with and without sensor adaptation. We have simulated three sensor characteristics to cover different sensor profiles with decreased visual correlation with respect to the original sensor. Our first simulated sensor consisted of a grayscale-only version of the data. We refer to it as ``grayscale sensor''. For some sensors, only panchromatic-EO (grayscale) imagery is available, thus we wanted to have a model that simulates lack of RGB color data. The second simulated sensor consists of swapping the position of the color channels from \textit{RGB} to \textit{BRG} in order to get images that looked different from the original. In addition, we dropped every other pixel for each color channel to simulate resolution loss between the two sensors (original and simulated). We refer to this simulated sensor as ``BRG-type1''. Finally, our third simulated sensor, ``BRG-type2'', consisted of swapping each color channel from \textit{RGB} to \textit{BRG}, as we did for ``BRG-type1'' and then applying different downsample factors for each channel to simulate more extreme resolution differences. We only kept one out of every eight pixels for the blue channel, every other pixel for the red channel, and only kept one out of every four pixels in the green channel. We then use nearest neighbor interpolation to align the three channels. 
\begin{figure*}[h]
\begin{center}
%\fbox{\rule{0pt}{2in} \rule{.9\linewidth}{0pt}}
	 \includegraphics[scale=0.48]{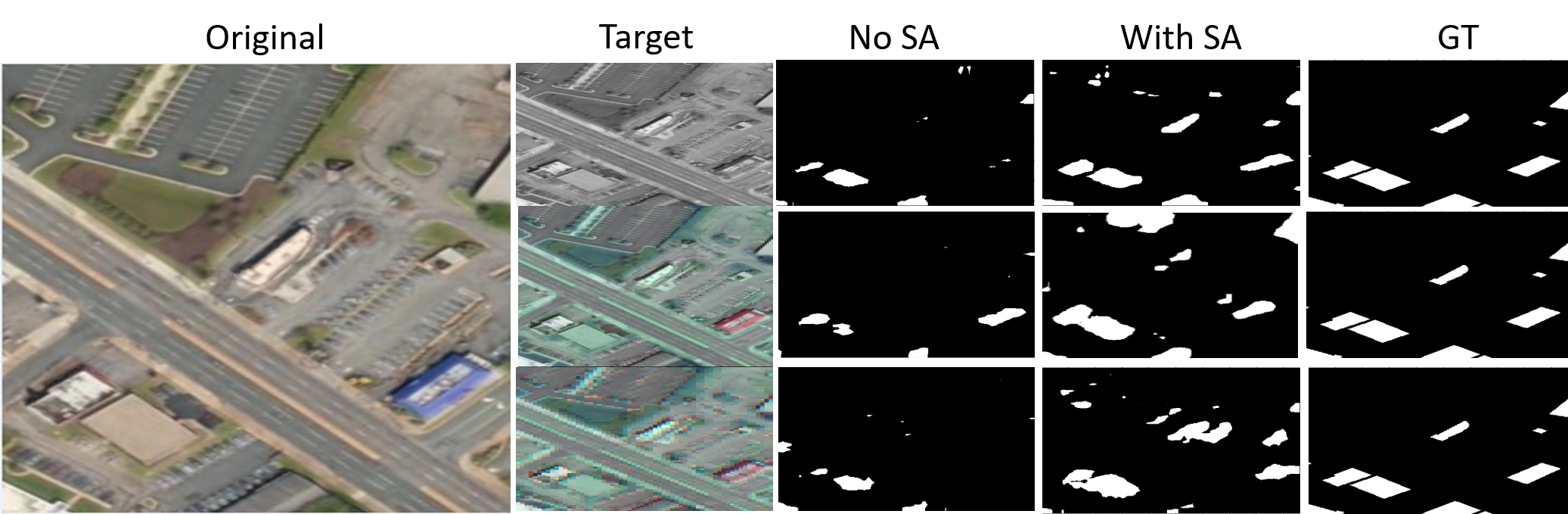}
   \caption{Sensor adaptation results. Left-most images show an example of the original scene/image (source sensor), second column shows the three versions of the original image simulated for each (target sensor), third column shows building segmentation results without using sensor adaptation, fourth column shows results with sensor adaptation during training, and fifth column shows the ground truth annotations for reference.}
\label{fig:sa_socom}
\end{center}
\end{figure*}
Figure~\ref{fig:sa_socom} shows a sample for the source sensor, the corresponding version for each of the three simulated target sensors for the same scene, and the building segmentation results for the segmenter with and without the translator module that performs the actual sensor adaptation.\\
\begin{table}[h]
\caption {Performance comparison between segmenter with and without SA for building segmentation applied on three simulated sensors (Grayscale, BRG-Type1, BRG-Type2).}
\begin{center}
  \begin{tabular}{ c | c | c }
    \hline
		\hline
    Sensor & mIoU & F1-score \\ 
		 %& Prec/Recall & Prec/Recall & Prec/Recall & Prec/Recall \\ 
		\hline
		\hline    
		Grayscale & 0.65 & 0.58 \\ 
		\hline
		Grayscale + SA & \textbf{0.76} & \textbf{0.68} \\ 
		\hline
    \hline
		BRG-type1 & 0.66 & 0.58 \\ 
		\hline
		BRG-type1 + SA & \textbf{0.73} & \textbf{0.64} \\ 
		\hline
    \hline
		BRG-type2 & 0.57 & 0.48 \\ 
		\hline
		BRG-type2 + SA & \textbf{0.68} & \textbf{0.58} \\ 
		\hline
    \hline
  \end{tabular}
\label{tab:sa_socom}
\end{center}
\end{table}
In Table~\ref{tab:sa_socom} we have included performance results of the segmenter (mIoU and F1-score) for both configurations -with and without SA. The test set contains more than 13000 buildings. In addition, in Table~\ref{tab:sa_socomana} we captured the performance gain of adding SA for building counts to quantify a more tangible benefit of SA. By running a simple analytics algorithm that counts the number of buildings in the scene, we compared both outputs to the ground truth (actual number of buildings).
\begin{table}[h]
\caption {Performance comparison between segmenter with and without SA for building counting task applied on three simulated sensors (Grayscale, BRG-Type1, BRG-Type2).}
\begin{center}
  \begin{tabular}{ c | c | c }
    \hline
		\hline
    Sensor & Building Counts & Difference \\
		 & & from Truth \\ 
		 %& Prec/Recall & Prec/Recall & Prec/Recall & Prec/Recall \\ 
		\hline
		\hline
    Ground truth & 13292 & 0 \\ 
		\hline
		\hline
		Grayscale & 14417 & +1125 \\ 
		\hline
		Grayscale + SA & 12975 & \textbf{-317} \\ 
		\hline
    \hline
		BRG-type1 & 16000 & +2708 \\ 
		\hline
		BRG-type1 + SA & 14709 & \textbf{+1417} \\ 
		\hline
    \hline
		BRG-type2 & 30819 & +17527 \\ 
		\hline
		BRG-type2 + SA & 16001 & \textbf{+2709} \\ 
		\hline
    \hline
  \end{tabular}
\label{tab:sa_socomana}
\end{center}
\end{table}

\subsection{Refined DeepLab and Source Adaptation Results}
\begin{figure*}[t]
\begin{center}
%\fbox{\rule{0pt}{2in} \rule{.9\linewidth}{0pt}}
	 \includegraphics[scale=0.58]{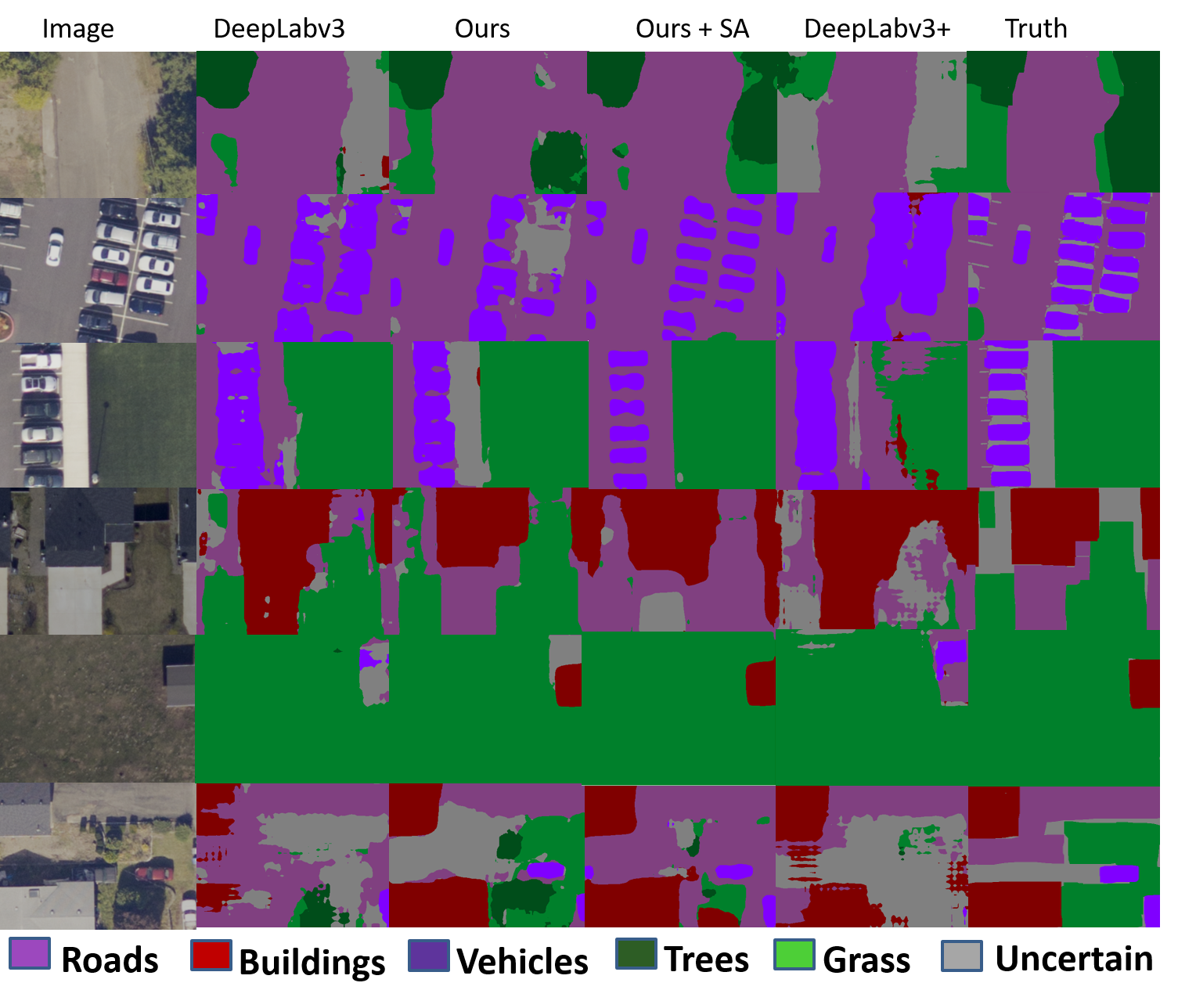}
   \caption{Comparison of results for several state-of-the-art models. Left column shows the original image chips, second column DeepLabv3/ResNet101 baseline results, third column our proposed improvements without source adaptation, fourth column shows outputs of our full model including source adaptation during training, fifth column shows DeepLabV3+ with MobileNet backbone, and right-most column shows the ground truth annotations for reference.}
\label{fig:comparison}
\end{center}
\end{figure*}

In this section we report our results after combining our proposed architecture and sensor adaptation. Table~\ref{tab:comp} shows mIoU and F1-score results for our proposed method with and without SA, as well as the original DeepLabv3/ResNet101 baseline framework. In addition, we added the recently proposed \textit{DeepLabv3+} with MobileNet as its backbone~\cite{deeplabv3plus}. Table~\ref{tab:compclass} presents the per-class (buildings, vehicles, roads) results. %RefineNet is an encoder-decoder architecture, with a ResNet101 encoder and a multi-path refinement decoder.
Finally, Figure~\ref{fig:comparison} shows visual results on overhead imagery samples for all algorithms.
\begin{table}[h]
\caption {Segmentation performance comparison (mIoU, F1-score).}
\begin{center}
  \begin{tabular}{ c | c | c }
    \hline
		\hline
    Algorithm & mIoU & F1-score \\ 
		 %& Prec/Recall & Prec/Recall & Prec/Recall & Prec/Recall \\ 
		\hline
		\hline
    DeepLabv3~\cite{deeplabv3} & 0.42 & 0.37 \\ 				
		\hline    
		Our Method & 0.43 & 0.45 \\ 
		\hline
		Our Method + SA & \textbf{0.51} & \textbf{0.47}\\ 
		\hline		
		DeepLabv3+~\cite{deeplabv3plus}& 0.44 & 0.39 \\ 
		(MobileNet) & & \\
		\hline
    \hline
  \end{tabular}
\label{tab:comp}

\end{center}
\end{table}

\begin{table*}[ht]
\caption {Segmentation performance comparison (mIoU, F1-score) for several classes.}
\begin{center}
  \begin{tabular}{ c | c | c | c | c | c | c }
    \hline
		\hline
    Algorithm & mIoU & mIoU & mIoU & F1-score & F1-score & F1-score \\ 
		 & vehicles & buildings & roads & vehicles & buildings & roads \\ 
		 %& Prec/Recall & Prec/Recall & Prec/Recall & Prec/Recall \\ 
		\hline
		\hline
    DeepLabv3~\cite{deeplabv3} & 0.32 & 0.34 & 0.43 & 0.36 & 0.31 & 0.41 \\ 				
		\hline    
		Our Method & 0.31 & 0.41 & \textbf{0.56} & 0.42 & 0.49 & \textbf{0.44}\\ 
		\hline
		Our Method + SA & \textbf{0.51} & \textbf{0.56} & 0.45 & \textbf{0.50} & \textbf{0.51} & 0.41 \\ 
		\hline		
		DeepLabv3+~\cite{deeplabv3plus} & 0.45 & 0.34 & 0.53 & 0.39 & 0.34 & 0.42 \\ 
		(MobileNet) & & \\
		\hline
    \hline
  \end{tabular}
\label{tab:compclass}

\end{center}
\end{table*}

%------------------------------------------------------------------------
\section{Discussion}
There are certain inherent challenges when processing images acquired with an airborne/spaceborne platform, including number of pixels that compose an object, separation space between objects (e.g., vehicles in parking lots), different viewing geometries, and geospatially diverse backgrounds. For these reasons, semantic segmentation from overhead imagery requires robust algorithms that can produce refined outputs. Our goal is to minimize over/under-segmentation as much as possible.\\
Another challenge is the lack of dense annotated data available from overhead platforms. Therefore, there is a need to recycle the current existing datasets. Being able to adapt existing datasets to different sensor platforms with different noise patterns, altitudes, and resolutions is key to being able to transfer current techniques applied to ground-based imagery to overhead imagery applications. Thus, we have designed a system that can provide accurate object segmentation while being able to re-use data from different sensors.\\
From the results we have shown, our proposed method improves DeepLabv3 baseline by producing more refined and accurate results. In Figure~\ref{fig:dlvstest}, we have shown several examples of the added value of our refinement module. We show that there are fewer cases of over/under-segmentation compared to the baseline. Also, in terms of performance metrics (see Table~\ref{tab:comp}), our approach can obtain more accurate segmentations than the baseline.\\
Adding sensor adaptation generated data allows for a larger improvement. Table~\ref{tab:comp} shows how we can boost the overall performance by significant differences. For instance, mIoU value jumps from $0.43$ to $0.51$.\\
When including other state-of-the-art algorithms in our comparisons, we observe that our proposed method remains competitive (with and without SA). Looking at the breakdown per class, we can see that it outperforms the other algorithms. See Table~\ref{tab:compclass} for more details.\\
Finally, Figure~\ref{fig:comparison} shows qualitative results of all compared algorithms, and we observe that our proposed methods is more visually similar to the ground truth.\\% Although SA seems to be hurting the road category (see rows 4 and 6 of Figure~\ref{fig:comparison}).\\
In terms of source adaptation experimentation, we aimed to study its limitations by applying different (simulated) sensor types. This allowed us to exercise the network to increasingly more distorted sensor models and observe the impact on semantic segmentation performance. Figure~\ref{fig:sa_socom} shows one example for all three types of sensors and how (fifth column) results with SA are much better than without. Even for the sensor \textit{BRG-type1} and \textit{BRG-type2} (color channel and resolution changes), the algorithm can still segment many buildings. The improvements are also noticeable when inspecting mIoU and F1-score results (Table~\ref{tab:sa_socom}). For all three sensor types, the segmentation algorithm using SA outperforms the non-SA version. In Tables~\ref{tab:comp} and ~\ref{tab:compclass} we see that the trend remains when applied to other (real) sensors in all but one class (roads). SA results show improvement with respect to the same system without the ``translator'' module. In addition, we performed a small exercise to translate the improvements to a more tangible outcome: building counting. For some applications, including urban planning and monitoring, population growth, or illegal land usage, building counts can offer an estimate of such activities. In order to count buildings, we counted connected components in the building mask from the segmentation and compared them to the manually annotated ground truth. For all three sensor types, our system with SA reported a much more accurate building count compared to the same algorithm without SA. In the worst case scenario, we were off by roughly 20\% (sensor \textit{BRG-type2}), while its counterpart without SA was off my more than 200\% for the same sensor. This exercise was performed to demonstrate the added value of accurate segmentations even when using images from a very different target sensor. See Table~\ref{tab:sa_socomana} for more details.\\
During our experimentation with GANs for source adaptation, we observed that the cycle loss was critical and that a low value for $\beta$ was failing to produce good ``translated'' images. This was somewhat expected, as the feature matching loss compares features extracted with an ImageNet pre-trained VGG19 network. A standard universal feature extractor does not efficiently capture visual attributes since there is little to no correlation between ground-based image features and those found in overhead imagery data.\\
As we enter an era of more ubiquitous imagery being available from overhead sensors, methods that can digest and process all visual data accurately will be necessary. Being able to leverage all the available datasets for overhead applications will enable much improved results. Performing feature transfer from objects observed from the ground to the overhead domain can help speed up the transition of current state-of-the-art systems to overhead applications. Exploring ground-to-space sensor/domain adaptation represents future work.

%------------------------------------------------------------------------
\section{Conclusion}
Given recent progress in semantic segmentation using deep learning, overhead imagery is an area that can leverage these advances and use it for improved image exploitation capabilities. In this work, we have expanded upon the DeepLabv3 architecture by adding a refinement stage that better adapts to overhead context. Our model aims to learn proper semantic partitions by performing progressive multi-resolution processing in the feature maps produced by the DeepLabv3 backbone. In addition, we have added a ``translator'' model based on cycle consistency adversarial domain translation to transfer the visual characteristics of images from a source sensor to images acquired by a target sensor where the system needs to be deployed. Through a series of experiments we showed how the proposed refinement blocks help the segmenter obtain refined and more accurate semantic maps. We have also shown how sensor adaptation can further improve segmentation performance by training on data with similar visual characteristics to the target sensor. Finally, we have shown how the combination of the refinement steps and sensor adaptation leads to improved results for the task of semantic segmentation of overhead imagery.

{\small
\bibliographystyle{ieee}
\bibliography{egbib}
}

\end{document}